\documentclass[letterpaper]{article} 
\usepackage{aaai24}  
\usepackage{times}  
\usepackage{helvet}  
\usepackage{courier}  
\usepackage[hyphens]{url}  
\usepackage{graphicx} 
\graphicspath{ {./figs/} }
\usepackage{natbib}  
\usepackage{caption} 
\usepackage{subcaption}
\usepackage{multirow}
\usepackage{amsmath}
\usepackage{amssymb}
\usepackage{booktabs}
\usepackage{xcolor}
\usepackage{lipsum}
\usepackage{algorithm}
\usepackage{algorithmic}

\usepackage{newfloat}
\usepackage{listings}
\DeclareCaptionStyle{ruled}{labelfont=normalfont,labelsep=colon,strut=off} 
\floatstyle{ruled}
\newfloat{listing}{tb}{lst}{}
\floatname{listing}{Listing}
\title{Box2Poly: Memory-Efficient Polygon Prediction of Arbitrarily Shaped \\ and Rotated Text}
\author {
    Xuyang Chen\textsuperscript{\rm 1,2},
    Dong Wang\textsuperscript{\rm 1}\thanks{Project Lead},
    Konrad Schindler\textsuperscript{\rm 3},
    Mingwei Sun\textsuperscript{\rm 1,4},\\
    Yongliang Wang\textsuperscript{\rm 1},
    Nicolo Savioli\textsuperscript{\rm 1},
    Liqiu Meng\textsuperscript{\rm 2}
}
\affiliations {
    \textsuperscript{\rm 1}Riemann Lab, Huawei\\
    \textsuperscript{\rm 2}Technical University of Munich\\
    \textsuperscript{\rm 3}ETH Zurich\\
    \textsuperscript{\rm 4}Wuhan University\\
    \{xuyang.chen, wangdong136, sunmingwei2, wangyongliang775, nicolo.savioli\}@huawei.com,schindler@ethz.ch,liqiu.meng@tum.de
}

\begin{document}

\maketitle

\begin{abstract}
Recently, Transformer-based text detection techniques have sought to predict polygons by encoding the coordinates of individual boundary vertices using distinct query features. However, this approach incurs a significant memory overhead and struggles to effectively capture the intricate relationships between vertices belonging to the same instance. Consequently, irregular text layouts often lead to the prediction of outlined vertices, diminishing the quality of results. To address these challenges, we present an innovative approach rooted in Sparse R-CNN: a cascade decoding pipeline for polygon prediction. Our method ensures precision by iteratively refining polygon predictions, considering both the scale and location of preceding results. Leveraging this stabilized regression pipeline, even employing just a single feature vector to guide polygon instance regression yields promising detection results. Simultaneously, the leverage of instance-level feature proposal substantially enhances memory efficiency ($>$ 50\% less vs.\ the SOTA method DPText-DETR) and reduces inference speed ($>$ 40\% less  vs.\ DPText-DETR) with comparable performance on benchmarks. The code is available at \url{https://github.com/Albertchen98/Box2Poly.git}.
\end{abstract}

\section{Introduction}

In an increasingly digitized world, the ability to automatically detect and extract textual information from images is an indispensable component for many applications of machine vision, including autonomous driving \cite{zhu2017cascaded,sun2022center}, SLAM \cite{li2020textslam}, Visual Place Recognition \cite{hong2019textplace}, and Assisted Navigation \cite{rong2016guided}. A reliable text detector is essential for localizing or parsing written text within a scene.
\begin{figure}[t!]
    \flushright
    \centering
    \subcaptionbox{}{\includegraphics[width=0.4\textwidth]{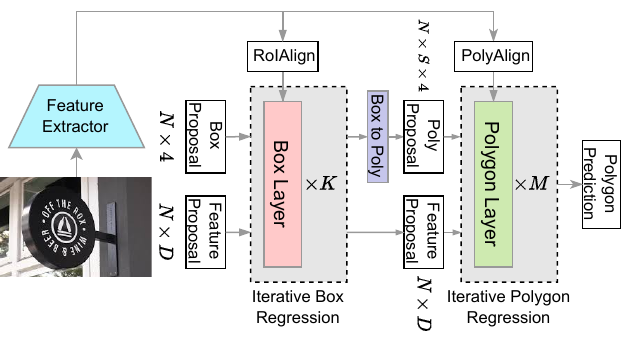}}
    \subcaptionbox{}{\includegraphics[width=0.3\textwidth]{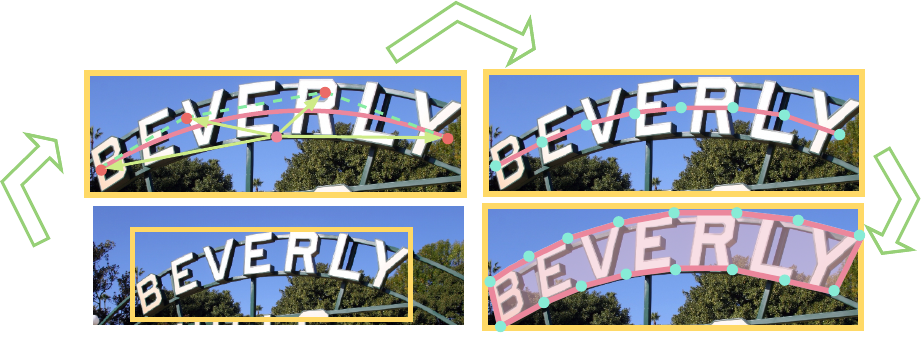}}
    \caption{(a) The overview structure of Box2Poly. (b) Transformation from Box to Poly.}
    \label{fig:box2poly_thumbnail}
\end{figure}
Recent text detection methods have diversified the regression targets to adapt to the diverse appearance of text instances. This variability typically arises in terms of text orientation and curvature level. 

Text orientation encompasses the possibilities of horizontal alignment or rotation, while text curvature spans the spectrum from straight to intricately bent. When dealing with text detection in cases where the layout is primarily straight and orientation is horizontal, a viable approximation of the effects of perspective projection caused by pinhole cameras is achievable through rotating and scaling bounding boxes \cite{zhang2016multi,ma2018arbitrary,he2021most} or utilizing quadrilaterals \cite{liu2017deep,bi2021disentangled}. 

Besides, for text instances with arbitrary shapes, where rotation and curvature can be present in a variety of configurations, the use of mask-based representations and contour-based methods has gained prominence due to their heightened adaptability. 

Mask-based approaches \cite{baek2019character,wang2019efficient,xu2019textfield,long2018textsnake} initially generate predictions at the pixel level, subsequently assembling individual pixels to construct segmentation masks at the instance level. Because masks possess nature at encapsulating deformable objects, these techniques offer a versatile solution for detecting text with irregular shapes. Nonetheless, their practical implementation is restricted by their substantial memory requirements and computationally intensive post-processing steps. Alternative approaches \cite{lyu2018mask,huang2022swintextspotter,zhang2019look} approach the detection task by framing it as instance segmentation. They incorporate a mask prediction head to generate segmentation masks from RoI(Region of Interest) proposals, which helps mitigate the computational workload to a certain degree, but the high-dimensional masks are hard to be effectively learned. 

On the other side, contour-based methods directly regress predictions towards the ground truth contours. Their effectiveness is notably impacted by the initial quality of the polygon proposals acquired. Therefore, several approaches \cite{dai2021progressive,zhang2023arbitrary} further explore mask-based techniques through initializing contour proposals from binary segmentation masks, followed by contour refinement. In a different vein, several methods \cite{ye2023dptext,zhang2022text} demonstrates a novel perspective by utilizing bounding boxes as the initial estimation of contours, either implicitly or explicitly. However, achieving a seamless transformation amidst the shape disparity between the box and the polygon contour remains an open challenge. Simultaneously, the memory efficiency of point query embedding is a notable concern.

In this paper, we delve deeper into the \textit{box-to-polygon} pipeline, focusing on addressing these issues.

As aforementioned, shape inconsistency arises during the box-to-polygon transformation process. This is especially evident when polygon vertices are formed by sampling points on boundaries \cite{ye2023dptext}. As a result, curved text instances that rotate more than 45 degrees incur an additional learning cost due to the misaligned initialization. A distinctive proposition introduced by DeepSolo \cite{ye2023deepsolo} involves utilizing top-\emph{K} Bézier curves \cite{liu2020abcnet} as initial proposals for the regression of the center polyline in text instances. While this approach enhances the suitability of the proposals for arbitrary shapes, it overlooks scale priors when generating Bézier curves across different feature map levels. Simultaneously, rather than refining the boundary directly, the center polyline undergoes recursive refinement during coordinate decoding. This makes the predictions more inclined to failure cases, such as self-intersection and misalignment to ground truth.  We address these concerns by generating polygon proposal from bounding box using Bézier centerline as intermediary.  As illustrated in Fig.~\ref{fig:box2poly_thumbnail}(b), given a text instance enclosed by a bounding box, a Bézier curve is generated within and expanded to match the same scale as the box. Subsequently, a polyline is derived by sampling a fixed number of vertices from this intermediate representation. Finally, we expand the polyline in its orthogonal direction to create a polygon, which participates into subsequent iterative regression. In this approach, the scale of polygon is inherited from the bounding box, while the Bézier centerline guarantees the flexibility of the orientation and shape.

The memory footprint overhead can be mitigated in two perspectives. First, unlike two-stage DETR paradigm \cite{ye2023dptext,zhang2022text,ye2023deepsolo}, which generates Top-\emph{K} proposals on the output of transformer encoder, we opt for a sparse proposal initialization \cite{sun2021sparse} to alleviate computational demands and reduce memory usage. This sparse proposal initialization involves introducing a fixed set of instances at the outset of coordinate decoding. As aforementioned, directly initializing text polygons leads to instability and complications, we utilize learnable bounding boxes as their prior estimates. We initiate the process by regressing the box proposals before proceeding to learn the polygon representations (refer to Fig.~\ref{fig:box2poly_thumbnail}). Furthermore, we employ instance-level feature embedding\footnote{corresponding to the term \emph{object query} in DETR \cite{carion2020end} and \emph{proposal feature} in Sparse RCNN \cite{sun2021sparse}} for each instance proposal. In contrast, TESTR \cite{zhang2022text} and DPText-DETR \cite{ye2023dptext} utilize point-level feature embedding for individual vertices, encoding their coordinates and interrelationships within each polygon instance.

Undoubtedly, instance-level feature embedding lacks the representative capacity when contrasted with point-level one. In light of this, our approach adopts the iterative regression strategy \cite{cai2018cascade,sun2021sparse} to ensure the stable regression of polygon coordinates. Diverging from refining with instance-independent absolute offsets \cite{ye2023dptext}, the predicted refinement offset in our method is made to be scale- and translation-invariant. 

Orientation discontinuity exists when the predicted polygon possesses reversed orientation w.r.t.\ ground truth, this hurdles the learning process since the network needs extra steps to correct the "wrong" orientation.  To accelerate the convergence, we augment the ground truth by extending the orientation-including annotation to be orientation-equivalent.

Sparse R-CNN \cite{sun2021sparse} is a representative work depending on iterative regression that achieves accurate results and high data-efficiency in object detection. At each decoding layer, RoIAlign \cite{he2017mask} is leveraged to extract image RoI feature for each box proposal. To conduct the task of polygon prediction, we propose a RoIAlign-like operation PolyAlign to assist the regression of polygons. 

In summary, the objectives of this paper are as follows:
\begin{itemize}
    \item Propose an \textbf{memory efficient} and \textbf{computationally efficient} text polygon prediction method with \textbf{comparable performance to SOTA}.
    \item Thoughtfully devise the \textbf{smooth transformation from box to polygon}.
    \item Introduce RoIAlign-like operation to extract \textbf{precise RoI features for polygon regression}.
    \item Stabilize the training process by \textbf{recursively refining polygon proposals}.
\end{itemize}
\section{Related Works}
\subsection{Iterative Regression Methods in Object Detection}
The iterative regression technique has garnered significant attention within the realm of bounding box object detection. Cascade R-CNN\cite{cai2018cascade} extends the architecture of Faster R-CNN \cite{ren2015faster} by introducing multi-stage detection sub-networks. This breaks down the regression task into a series of cascade layers, where each layer's regressor operates directly on the prediction results provided by the previous layer. This sequential refinement process contributes to an improvement in detection quality. Similarly, within the framework of DETR methods, such as Deformable DETR \cite{zhu2020deformable} and DAB DETR \cite{dai2021dynamic}, a comparable strategy is employed. These methods also embrace iterative bounding refinement, where the regression head of each decoder layer predicts bounding boxes as relative offsets w.r.t\ the bounding boxes of the preceding layer. This iterative approach contributes to the gradual enhancement of localization accuracy. In a different vein, Sparse R-CNN \cite{sun2021sparse} introduces an innovative paradigm by circumventing the computationally intensive proposal generator in the Fast R-CNN paradigm. Instead, they initialize learnable bounding boxes with thoroughly sparse setting and subsequently update them using a methodology akin to that of Cascade R-CNN. Despite its streamlined pipeline, this approach achieves remarkable performance and training efficiency.
\begin{figure}[t]
 \centering
 \includegraphics[width=0.4\textwidth]{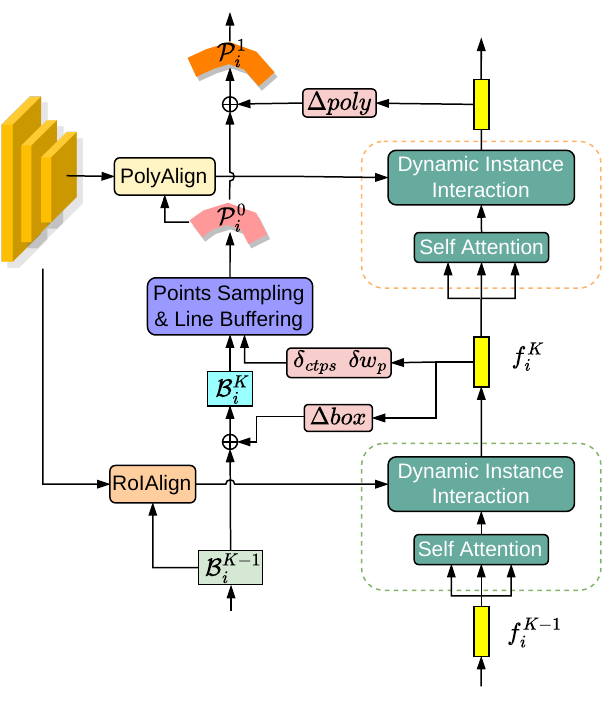}
 \caption{Illustration of the last bounding box regression iteration and the first polygon regression iteration.}
 \label{fig:box2bezier2poly_regression}
\end{figure}

\subsection{Text Detection With Contour Iterative Regression}
Recognizing the complexity of accomplishing the regression in a single pass, researchers have explored the application of iterative regression to predict concise contour boundaries for text instances,  An example is TextBPN++ \cite{zhang2023arbitrary}, where the boundary originates from a pixel-level score map and is subsequently refined through recursive optimization facilitated by a boundary transformer. However, this transformation from a score map to a boundary is heavily reliant on predefined rules and the pixel-level prediction restricts its application to higher input resolution.
Leveraging on the impressive performance of DETR methods, TESTR \cite{zhang2022text} takes a novel approach rooted in Deformable DETR \cite{zhu2020deformable}. It harnesses bounding boxes predicted by the transformer encoder's output as proposals and then transforms these boxes into polygons by sampling points along the upper and lower sides. Subsequently, the generated polygon vertex is embedded into point query through sine positional encoding and participates into coordinate regression at each layer of the decoder. A further advancement, DPText-DETR \cite{ye2023dptext}, refines the TESTR framework by introducing iterative refinement of polygon vertices during decoding. It also introduces the utilization of vertex coordinates as reference points for conducting deformable cross-attention \cite{zhu2020deformable}, in response to each point query.
DPText-DETR surpasses the performance of TESTR and emerges as the new state-of-the-art method across several text detection benchmarks. However, DPText-DETR shares a common challenge with TESTR, namely, a substantial memory footprint. This is attributed to the requirement of encoding coordinate information for each polygon vertex with an individual query.
\section{Method}
\label{sec:method}
\subsection{Overview}
The overall structure of Box2poly Network is illustrated in Fig.~\ref{fig:box2poly_thumbnail}(a), which is built upon Sparse R-CNN \cite{sun2021sparse}. The image containing text instance is fed into an FPN backbone to extract multiple layer features. Then the decoding part on the right can leverage these features to complete the detection task. Like in Sparse R-CNN, a fixed, learnable set of $N$ bounding boxes serve as region proposals, and each of those proposals is corresponding to a feature proposal of dimension $D$. The network performs structured regression with two heads, one for boxes and one for polygons. Each of them contains $K$ and $M$ layers respectively. The box proposals serve as priors for the subsequent polygon prediction. To transition from box prediction to polygon representation, a transitional layer \emph{Box to Poly} is inserted between the last layer of box head and the first layer of polygon head. PolyAlign is leveraged to extract precise RoI features for polygon regression. Besides that, orientation-equivalent annotation for text polygon is introduced to eliminate the learning discontinuity when the predicted polygon shows reversed orientation w.r.t.\ the spatially proximal ground truth one. 
\subsection{Learnable Bounding Boxes as Priors}
\label{sec:box_priors}
We leverage learnable bounding boxes to estimate the location and scale of text instances before polygon regression. 
Following Sparse RCNN \cite{sun2021sparse}, a set of learnable bounding boxes are initialized with the image size and refined recursively for \(K\) times. As illustrated in Fig.~\ref{fig:box2bezier2poly_regression}, given a box proposal \(\mathcal{B}_i^{K-1},i \in N\) from previous layer, RoIAlign \cite{he2017mask} is leveraged to extract RoI feature within its scope. This extracted RoI feature is merged with its corresponding proposal feature \(f_i^{K-1}\) using Dynamic Instance Interaction \cite{sun2021sparse}, yielding an updated proposal feature \(f_i^{K}\). To further refine the bounding box proposal \(\mathcal{B}_i^{K-1}\), a coordinate offset \(\Delta box\) is predicted based on \(f_i^{K}\) to achieve \(\mathcal{B}_i^{K}\).
\begin{figure}[t]
    \centering
    \includegraphics[width=0.45\textwidth]{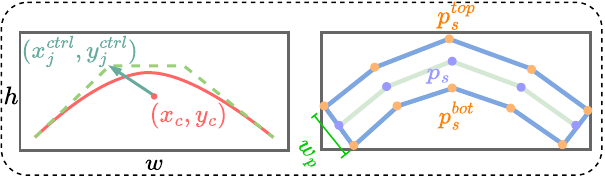}
    \caption{Ilustration of transforming from box to polygon with Bézier curve as intermediate representation.}
    \label{fig:polygon_from_bezier_math}
\end{figure}
\subsection{Box to Poly}
\label{sec:box2poly}
To conduct polygon regression, we first transform the bounding box prior \(\mathcal{B}_i^{K}\) to polygon representation using Bézier curve \cite{liu2020abcnet} as intermediary. As depicted in Fig.~\ref{fig:box2bezier2poly_regression} and Fig.~\ref{fig:polygon_from_bezier_math}, at the final layer \(\left(K-1\right)\) of iterative bounding box regression, a Bézier curve with four control points is predicted upon proposal feature \(f_i^{K}\) as relative offsets \([\delta p_{j} = \left(\delta x_{j}, \delta y_{j}\right)]_{j=0}^3\) w.r.t.\ the center of the reference box. We denote reference bounding box as  \(\left(x_{c},y_{c},w,h\right) \), where \(\left(x_{c},y_{c}\right) \) represents the box center and \(\left(w,h\right) \) refers to the box scale. The control points \(\left(x^{ctrl}_{j}, y^{ctrl}_{j}\right)\) of the cubic Bézier curve are expressed as
\[\left(x^{ctrl}_{j}, y^{ctrl}_{j}\right) = \left(x_{c},y_{c}\right) + \left(\delta x_{j} \cdot w, \delta y_{j} \cdot h\right).\\
\]
By multiplying \(\left(\delta x_{j}, \delta y_{j}\right)\) with \(\left(w, h\right)\), the generated Bézier curve inherits the scale from the reference bounding box. Next, we uniformly sample points on this Bézier curve to obtain vertices of a polyline that fits the center-curve of text instance. This process is performed using the following equations:
\begin{flalign*}
& p_{s,x} = \sum_{j=0}^{3} x^{ctrl}_{j} B_{j}(t_s),\;\;
p_{s,y} =  \sum_{j=0}^{3} y^{ctrl}_{j} B_{j}(t_s),\\
& B_{j}(t_s) = \binom{3}{j} t_s^j (1-t_s)^{3-j},\;
t_s = \frac{s}{S-1},
\end{flalign*}
where \(s\in\left[0,\hdots,S-1\right]\) indexes each sample point, and \(\left(p_{s,x}, p_{s,y}\right)\) represents their coordinates. \(B_{j}(t_s)\) calculates the value of each Berstein basis polynomial \cite{lorentz2012bernstein} at every single step \(t_s\). 
The polyline is complemented with a scalar width to form a boundary around a text instance. To that end an additional value \(\delta w_{p}\) is predicted, which describes the scaling factor between the bounding box size and the initial value of the text polygon's width:
\[w_{p} = \sqrt{wh} \cdot \exp(\delta w_{p}).\]
To obtain polygon vertices, we expand the polyline with distance \(w_{p}/2\) towards outside using \(f_{buffer}\):
\[
p_s^{top}, p_s^{bot} = f_{buffer} (p_s,\frac{w_p}{2}).
\]

The complete procedure is depicted in Figure~\ref{fig:polygon_from_bezier_math} to enhance comprehension.
To ensure a distinctive representation, the \emph{top} vertices \(p_s^{top}\) are designated to reside on the left side while traversing the polyline in vertex order.
\subsection{Iterative Regression of Proposal Polygons}
\label{sec:poly_regression}
A set of polygon proposals \(N \times S \times 4\) are generated from last step after transforming \(N\) proposal boxes. Similar to the proposal boxes, their coordinates undergo updates across \(M\) layers of iterative regression. For instance, as depicted in Figure~\ref{fig:box2bezier2poly_regression}, during the initial polygon regression layer, given a proposal \(\mathcal{P}_i^{m-1}=\left[\left(p_{s,y}^{top}, p_{s,y}^{bot}\right)^{m-1}\right]_{s=0}^{S-1}\) from the preceding layer, we first apply the following transformation:
\[
\begin{split}
&\left(p_{s,x}, p_{s,y}, p_{s,dx}, p_{s,dy}\right)^{m-1} = \\
&\left(\frac{p_{s,x}^{top} + p_{s,x}^{bot}}{2}, \frac{p_{s,y}^{top} + p_{s,y}^{bot}}{2},p_{s,x}^{top} - p_{s,x}^{bot}, p_{s,y}^{top} - p_{s,y}^{bot}\right)^{m-1}
\end{split}
\]
This transformation converts the representation to center point coordinates and the coordinate differences between the top and bottom vertices.
To ensure stability in the iterative regression process, the polygon regression head yields an output denoted as a scale- and location-invariant distance vector \(\Delta_{poly} = \left[\left(\delta p_{s,x},\delta p_{s,y},\delta p_{s,dx}, \delta p_{s,dy} \right)\right]_{s=0}^{S-1}\).
Subsequently, the \(m^\text{th}\) decoder refines the polygon as follows:
\begin{align*}
p_{s,x}^m & = p_{s,x}^{m-1} + \delta p_{s,x} \lvert p_{s,dx}^{m-1} \rvert,\\
p_{s,y}^m & = p_{s,y}^{m-1}  + \delta p_{s,y} \lvert p_{s,dy}^{m-1} \rvert,\\
p_{s,dx}^m & = p_{s,dx}^{m-1}  \exp{\left(\delta p_{s,dx} \right)},\\
p_{s,dy}^m & = p_{s,dy}^{m-1}  \exp{\left(\delta p_{s,dy} \right)}.
\end{align*}
Gradients are soly back-propagated through distance vector \(\Delta_{poly}\) to make the training stable, as in \cite{zhu2020deformable,sun2021sparse}.
Afterwards, \(\left(p_{s,x}, p_{s,y}, p_{s,dx}, p_{s,dy}\right)^m\) are transformed back to the representation \(\mathcal{P}_i^{m}=\left[\left(p_{s,y}^{top}, p_{s,y}^{bot}\right)^m\right]_{s=0}^{S-1}\) to align with the ground truth and compute the coordinate loss.

As \(\exp{\left(\delta p_{s,dx}\right)}\) and \(\exp{\left(\delta p_{s,dy}\right)}\) are invariably greater than 0, the sign of \(\left(p_{s,dx},p_{s,dy}\right)^{m}\) remains consistent with \(\left(p_{s,dx},p_{s,dy}\right)^{m-1}\). In simpler terms, this ensures that the updated polygon maintains a coherent alignment of its top and bottom boundaries across various regression layers.

\subsection{PolyAlign}
\label{sec:polyalign}
In pursuit of more representative Region of Interest (RoI) features for each polygon proposal, we introduce an innovative step called PolyAlign. In contrast to the conventional bounding box RoIAlign \cite{he2017mask}, which employs bilinear interpolation at the center of each RoI grid, PolyAlign directly operates on the paired top and bottom vertices of polygons and their corresponding centerline vertices. This approach is underpinned by the assumption that point-level coordinate regression is especially sensitive to deviations in the foothold of RoI feature extraction.

Upon extracting Poly RoI features \([\boldsymbol{z}^1,\boldsymbol{z}^2..,\boldsymbol{z}^L]\) from multi-layered feature maps, we aggregate them using sum aggregation \(\boldsymbol{z}^{ms}=\sum_{l=1}^{L} \boldsymbol{z}^l\). Here, \(L\) refers to the number of multi-layer feature maps and \(\boldsymbol{z}^{ms}\) represents the aggregated feature. The resultant Polygon RoI feature assumes spatial dimensions of \(S \times 3 \times 256\), where \(S\) is the number of vertices on each boundary.
\begin{figure}[!htb]
    \centering
    \includegraphics[width=0.45\textwidth]{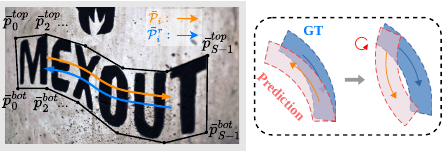}
    \caption{(left) The extended polygon annotations. \(\mathcal{P}_i\) represents the original order, while \(\mathcal{P}_i^r\) is the reversed one. (right) Learning orientation reversal under disparity.}
    \label{fig:reverse_gt}
\end{figure}
\subsection{Orientation-Equivalent Annotation}
\label{sec:oritation-quavalient-anno}
As the polygon representation, with pair-wise top and bottom vertices, always carries the attribute of orientation, an issue similar to the boundary discontinuity \cite{yang2021learning,yang2021rethinking,yang2022kfiou} exists in its regression. As illustrated in Fig.~\ref{fig:reverse_gt} (right), if the polygon is initialized with reverse orientation w.r.t.\ the ground truth, the network needs extra effort to learn the reversal meanwhile correcting the vertex coordinates, which greatly complicates the learning process in the presence of diverse rotations.
Inspired by \cite{liao2022maptr}, we remedy this issue by augmenting the ground truth with its orientation-reversed version. Given the ground truth as \(\Bar{\mathcal{P}}_i=\left[ \left( \Bar{p}_0^{top},\Bar{p}_0^{bot} \right),...,\left( \Bar{p}_{S-1}^{top},\Bar{p}_{S-1}^{bot} \right) \right]\) in Fig.~\ref{fig:reverse_gt} (left),  the reversed version is then \(\Bar{\mathcal{P}}_i^r=\left[ \left( \Bar{p}_{S-1}^{bot},\Bar{p}_{S-1}^{top} \right),..., \left( \Bar{p}_0^{bot},\Bar{p}_0^{top} \right) \right]\). Given both orientations, a candidate polygon can, during training, be regressed to either \(\Bar{\mathcal{P}}_i\) or \(\Bar{\mathcal{P}}_i^r\). 
\subsection{Optimization}
\subsubsection{Polygon Bipartite Matching}
Like other detectors \cite{carion2020end,sun2021sparse,dai2021dynamic,zhu2020deformable} that perform set prediction, we employ Bipartite Matching to assign predictions to ground truth.
This cost consists of two components: correctness of class prediction and  distance between coordinates. Given prediction set \(Y=\{y_i\}_{i=0}^{N-1}\) and padded ground truth set \(\Bar{Y}=\{\Bar{y}_i\}_{i=0}^{N-1}\). The matching cost is defined as 
\begin{align*}
\mathcal{L}^{poly}_{match} \left(y_{\pi(i)}, \Bar{y}_i\right) = 
\mathcal{L}_{focal} \left(\rho_{\pi(i)}, \Bar{c}_i\right) + \mathcal{L}_{poly} \left(\mathcal{P}_{\pi(i)}, \Bar{\mathcal{P}}_i^E\right).
\end{align*}
Here, \(\pi_{(i)}\in\Pi_N\) represents a permutation of N elements. \(\mathcal{L}_{focal} \left(\rho_{\pi(i)}, \Bar{c}_i\right)\) evaluates the class matching cost with Focal loss \cite{lin2017focal} between the predicted class score \(\rho_{\pi(i)}\) and ground truth label \(\Bar{c}_i\), while \(\mathcal{L}_{poly} \left(\mathcal{P}_{\pi(i)}, \Bar{\mathcal{P}}_i^E\right)\) calculates the coordinate matching cost for polygon vertex coordinates by \(L1\) distance between prediction \(\mathcal{P}_{\pi(i)}\) and extended ground truth \(\Bar{\mathcal{P}}_i^E\). \(\mathcal{L}_{poly} \left(\mathcal{P}_{\pi(i)}, \Bar{\mathcal{P}}_i^E\right)\) is made to be orientation-invariant by taking the minimum one between two prediction-target pairs : \(\mathcal{P}_{\pi(i)} \sim \Bar{\mathcal{P}}_i\) and \(\mathcal{P}_{\pi(i)} \sim \Bar{\mathcal{P}}^r_i\). This is expressed as
\begin{flalign*}
& \mathcal{L}_{poly}\left(\mathcal{P}_{\pi(i)}, \Bar{\mathcal{P}}^E_i\right) = \mathcal{L}_{poly}\left(\mathcal{P}_{\pi(i)}, \Tilde{\mathcal{P}}_i\right), \\
& \Tilde{\mathcal{P}}_i = \underset{\left[\Bar{\mathcal{P}}_i, \Bar{\mathcal{P}}^r_i\right]}{arg min} \left(\|\mathcal{P}_{\pi(i)}-\Bar{\mathcal{P}}_i\|_1, \|\mathcal{P}_{\pi(i)}-\Bar{\mathcal{P}}^r_i\|_1 \right).
\end{flalign*}
\subsubsection{Set Prediction Loss}
The training loss is identical to the matching cost, but summed only over the matched pairs. To encourage consistency, the ground truth for polygon coordinates directly inherits the orientation 
\(\Tilde{\mathcal{P}}_i\) from the matching cost calculation.
\begin{table*}\normalsize
\centering
\setlength{\tabcolsep}{4pt}
\resizebox{0.85\hsize}{!}{
\begin{tabular}{c c|ccc|ccc} 
\toprule[1.5pt]
\multirow{2}{*}{\textbf{Method}} & \multirow{2}{*}{\textbf{Feature Extractor}} & \multicolumn{3}{c|}{\textbf{TotalText}} & \multicolumn{3}{c}{\textbf{CTW1500}}\\
& & P & R & F & P & R & F \\
\midrule[1.1pt]
DB\cite{liao2020real} & ResNet50+DCN & 87.1 & 82.5 & 84.7 & 86.9 & 80.2 & 83.4 \\
\hline
I3CL\cite{du2022i3cl} & ResNet50+FPN & 89.2 & 83.7 & 86.3 & 87.4 & 84.5 & 85.9 \\
\hline
ABCNet-v2\cite{liu2021abcnet} & ResNet50+FPN & 90.2 & 84.1 & 87.0 & 85.6 & 83.8 & 84.7 \\
\hline
Boundary(end-to-end)\cite{wang2020all} & ResNet50+FPN & 88.9 & 85.0 & 87.0 & - & - & -\\
\hline
TESTR-Polygon\cite{zhang2022text} & ResNet50+Deform. Encoder & 93.4 & 81.4 & 86.9 & 92.0 & 82.6 & 87.1 \\ 
\hline
SwinTextSpotter\cite{huang2022swintextspotter} & Swin-Transformer+FPN & - & - & 88.0 & - & - & 88.0 \\ 
\hline
DPText-DETR\cite{ye2023dptext} & ResNet50+Deform. Encoder & 91.8 & 86.4 & \textbf{89} & 91.7 & 86.2 & \textbf{88.8} \\ 
\hline
Box2Poly(ours) & ResNet50+FPN & 90.22 & 86.57 & \underline{88.35}  & 88.84 & 87.46 & \underline{88.13} \\ 
\bottomrule[1.5pt]
\end{tabular}  
}
\caption{Quantitative text detection results on arbitrarily shaped datasets, measured using Precision (P), Recall (R), and F-score (F). Our method's results are presented as \textit{mean} achieved through five rounds of finetuning, the spread of F-score is conveyed as \emph{$3\sigma$ empirical standard deviation}, which are correspondingly $\pm$0.26 and $\pm$0.35 for TotalText and CTW1500.}
\label{tab:perf_compare}
\end{table*}
\section{Experiments}
\subsection{Datasets}
The datasets involved in the experiment are \textbf{SynthText 150K}, \textbf{TotalText}, \textbf{CTW1500}, \textbf{ICDAR19 MLT} and \textbf{InverseText}. As implied by the name, \textbf{SynthText 150K} \cite{liu2020abcnet} collects 150k synthesized scene text images, consisting of 94,723 images with multi-oriented straight texts and 54,327 images with curved ones. \textbf{TotalText} \cite{ch2017total} contains 1,255 training images and 300 test images with highly diversified orientations and curvatures. In this dataset, the text instance is labeled in word-level. \textbf{CTW1500} is another dataset with curved texts including 1,000 training images and 500 test images. Its annotation is in text-line level. \textbf{ICDAR19 MLT} is a multi-lingual scene text detection dataset \cite{nayef2019icdar2019} providing 10k images for training. 
\textbf{InverseText} \cite{ye2023dptext} is a test set that consists of 500 test images with text instances concurrently possessing rotation and curvature.

\subsection{Implementation Details}
ResNet50 \cite{he2016deep} is adopted for all experiments and initialized with ImageNet pretrained weights. The batch size is set 16 and all models are trained with 8 pieces NVIDIA RTX 3090 GPUs. The final results on TotalText and CTW1500 are reported with the training strategy similar to \cite{zhang2022text,ye2023dptext}: First, the network is pretrained on a combined dataset for 180k iterations with a learning rate \(2.5 \times 10^{-5}\) that drops at 144k and 162k step. The learning rate drop factor is 10. The combined dataset is composed of TotalText, ICDAR19 MLT and SynthText 150K. 
Following \cite{ye2023dptext}, data augmentation includes instance aware random cropping, random blur, color jittering and multi-scale resizing, with the shorter side restricted to the range $[480,\hdots, 832]$ pixels and the longer side capped at a maximum of 1600 pixels.

Before training, the datasets are polished following the protocol of \cite{ye2023dptext} -- annotation of polygon vertices is rearranged to eliminate the influence of the implicit reading order, and the training set is expanded by rotating training images with a set of predefined rotation angles.  Our chosen proposal number, denoted as \(N\), is set at 300. Meanwhile, for Bezier curves, the number of sampling points \(S\) has been established as 8, yielding 16 vertices for each polygon proposal. The box head employs a multi-stage approach with \(K\) layers set at 3, while the polygon head similarly employs a multi-stage structure with \(M\) layers also set at 3.
\begin{figure}[t]
 \centering
 \includegraphics[width=0.45\textwidth]{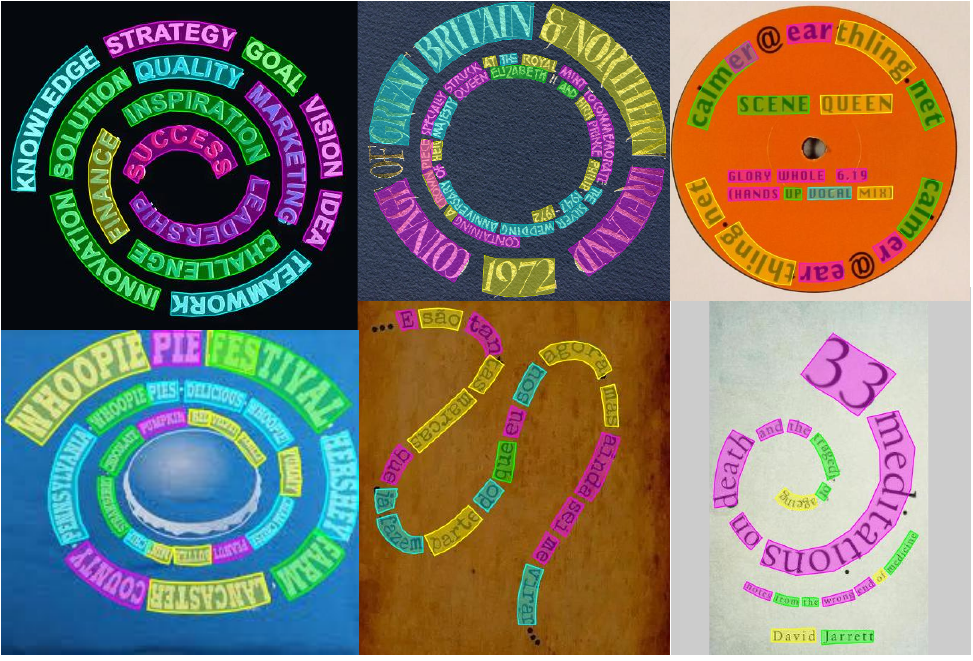}
 \caption{Qualitative results on images containing text instances rotated and randomly curved.}
 \label{fig:result_visualization}
\end{figure}

\subsection{Comparisons With Other Methods}
\subsubsection{On Arbitrary-Shaped Text Benchmarks}
We evaluate our network on two widely-used text detection benchmarks, namely TotalText and CTW1500. Despite using just one feature embedding per polygon instance, our approach achieves comparable performance to the current state-of-the-art method, DPText-DETR. The latter utilizes point-level feature embedding on each polygon vertex (16 vertices per polygon). In comparison, our method's F-Score is only 0.7\% lower on both dataset: 88.35 vs.\ 89 on TotalText and 88.13 vs.\ 88.8 on CTW1500 (Tab.~\ref{tab:perf_compare}). The visual comparison can be found in the Appendix.
Moreover, it is noteworthy that DPText-DETR employs a stronger feature extracting method compared to our network, and it takes advantage of proposal generator. Box2Poly leads TESTR-Polygon by 1.7\% and 1.2\% on these two benchmarks (Tab.~\ref{tab:perf_compare}).
Additionally, when compared to SwinTextSpotter, which employs a stronger backbone, our detection performance overhead on TotalText and CTW1500 is 0.35 and 0.13 regarding F-score, respectively (Tab.~\ref{tab:perf_compare}).
\subsubsection{Stability Against Rotation Together With Arbitrary Shape}
To showcase the robustness of our network in simultaneously managing text rotation and arbitrary curvature, we conduct a comparative analysis of detection results with other polygon-based methods on InverseText. As delineated in Tab.~\ref{tab:inversetext_results}, our network achieves marginally improved performance over DPText-DETR under the setting of no pretraining. With the inclusion of pretraining, Box2Poly achieves an F-score of 88.7, surpassing DPText-DETR by 1.6\%.
Further substantiating our network's capabilities, we visually present a selection of detection results in Fig.~\ref{fig:result_visualization}. Notably, the network consistently generates compact polygons for images featuring text instances with varying degrees of rotation and curvature.
\subsubsection{Memory Efficiency and Data Efficiency}
As depicted in Tab.~\ref{tab:no_pretrain_results}, the incorporation of the \emph{single feature embedding per instance} approach notably reduces the GPU memory footprint of our network compared to DPText-DETR, despite our model leveraging three times as many instance proposals (300 vs.\ 100). And a faster inference speed is observed in our method (FPS 19.6 vs.\ 13.6). Remarkably, after solely 10k training iterations on the TotalText dataset, Box2Poly achieves performance on par with no-pretraining DPText-DETR. This result underscores Box2Poly's commendable data efficiency(Tab.~\ref{tab:no_pretrain_results}).
\begin{table*}\normalsize
\centering
\setlength{\tabcolsep}{4pt}
\resizebox{0.9\hsize}{!}{
\begin{tabular}{c |c c|c c c|c|c|c c} 
\toprule[1.5pt]
\multirow{2}{*}{\textbf{Experiments}} & \multicolumn{2}{c|}{\textbf{Trans. Box to Poly}} & \multicolumn{3}{c|}{\textbf{PolyAlign}} & \textbf{Orientation-Equivalent} & \multirow{2}{*}{\textbf{Polygon IoU Loss}}  & \multicolumn{2}{c}{\textbf{F-score}}\\
& Sampling & Box+bezier & Vertex & Grid & Bezier & \textbf{Annotation} & & 50K & 30K\\
\midrule[1.1pt]
1. & \checkmark & - & \checkmark & - & - & \checkmark & - & 53.61 & - \\
2. & - & \checkmark & - & \checkmark & - & \checkmark & - & 86.07 & - \\
3. & - & \checkmark & - & - & \checkmark & \checkmark & - & 86.26 & - \\
4. & - & \checkmark & \checkmark & - & - & - & - & \textbf{87.49} & 86.94 \\
5. & - & \checkmark & \checkmark & - & - & \checkmark & \checkmark  & 86.29 & - \\
\hline
6.(final) & - & \checkmark & \checkmark & - & - & \checkmark & - & 87.41 & \textbf{87.34} \\
\bottomrule[1.5pt]
\end{tabular}
}
\caption{Ablation studies on InverseText. 50K and 30K corresponds to the number of training iterations. "final" corresponds to the version of model we leverage to derive the final result. \(\checkmark\) marks the integrated component or technique.}
\label{tab:ablations}
\end{table*}

\begin{table}[t]\normalsize
    \centering
    \setlength{\tabcolsep}{4pt}
    \resizebox{0.6\hsize}{!}{
    \begin{tabular}{c|c c c}
    \toprule[1.5pt]
    \textbf{Method} & P & R & F\\
    \midrule[1.1pt]
    TESTR-Polygon & 91.9 & 84.4 & 86.8 \\
    \hline
    DPText-DETR & 90.7 & 84.2 & 87.3\\
    \hline
    Box2Poly(ours) & 91.4 & 86.1 & \textbf{88.7}\\
    \hline
    \hline
    DPText-DETR\dag & - & - & 86.8\\
    \hline
    Box2Poly\dag(ours) & 90.4 & 85.7 & \textbf{87.41}\\
    \bottomrule[1.5pt]
    \end{tabular}
    }
    \caption{Evaluation of methods on the InverseText dataset. \dag means pretraining is not included. The results of the other two methods are fetched from DPText-DETR.}  
    \label{tab:inversetext_results}
\end{table}
\subsection{Ablation Studies}
In this section, we undertake an analysis of the distinct contributions made by various components introduced in Sec.~\ref{sec:method}. Our experimental approach involves systematically deactivating or substituting different components of Box2Poly during each experiment. The model is trained for 50k iterations on the TotalText dataset, with subsequent evaluation conducted on InverseText. The initial learning rate is set at \(2.5\times10^{-5}\) and is reduced to \(2.5\times10^{-6}\) at 80\% of the iterations. The confidence score threshold is established at 0.3. Comprehensive results are detailed in Tab.~\ref{tab:ablations}.
\subsubsection{Transformation From Box to Polygon}
In Experiment 1, as outlined in Tab.~\ref{tab:ablations}, we employ a strategy akin to DPText-DETR \cite{ye2023dptext} and TESTR \cite{zhang2022text} for the conversion from bounding box to polygon. This involves sampling a consistent number of points along the upper and lower edges of the bounding box, ensuring equidistant distribution. It's performance is significantly lower compared to our \emph{Box to Poly} method discussed in Sec.~\ref{sec:box2poly}. This verifies the effectiveness of our design conducting transformation.
\subsubsection{PolyAlign}
In order to validate the assumption introduced in Sec.~\ref{sec:polyalign}, we conduct an evaluation of the impact of different PolyAlign implementations: Vertex, Grid and Bézier. Among these, PolyAlign-Vertex is the implementation integrated into our proposed network. This variant projects vertex coordinates back onto the feature maps and performs RoI feature extraction from there. On the other hand, PolyAlign-Grid aligns its feature extraction foothold with the center of sampling grids within the polygon, following the design principles of RoI Align \cite{he2017mask}. The Bézier implementation, PolyAlign-Bézier, is based on BezierAlign from ABCNet \cite{liu2020abcnet}, albeit with several modifications made to ensure uniformly distributed sampling grids. 
To cope with this feature extractor, the regression is conducted on Bézier-Polygon, where top and bottom boundary is represented with Bézier curves. As indicated in Tab.~\ref{tab:ablations}, it is observed that PolyAlign-Vertex  achieves the best performance. This result lends support to our assumption that precise feature extraction at specific locations is advantageous for vertex regression. This finding is congruent with the notion that DPText-DETR outperforms TESTR-Polygon due to the former using vertex coordinates as reference points to conduct deformable cross-attention \cite{zhu2020deformable}, rather than the center of proposal bounding box.
\subsubsection{Orientation-Equivalent Annotation}
As demonstrated in the results of $50K$ training iterations presented in Tab.~\ref{tab:ablations}, the incorporation of \emph{OEA} (\emph{Orientation-Equivalent Annotation}) does not yield a performance improvement for the network. However, upon integrating \emph{OEA}, a better convergence behavior is observed after re-running Exp. 4 and 7 with identical configuration for less iterations($30K$) (Tab.~\ref{tab:ablations}). As a result, the disruption in learning caused by orientation discrepancy has been notably mitigated. The relatively modest impact of \emph{OEA} on performance can be attributed to the fact that, within the polished datasets, text orientation issue has been mitigated to a certain extent.

\begin{table}[t]
    \centering
    \resizebox{0.9\hsize}{!}{
    \setlength\tabcolsep{4pt}
    \begin{tabular}{c|c c c|c|c}
    \toprule[1.5pt]
    \multirow{2}{*}{\textbf{Method}}  & \multicolumn{3}{c|}{\textbf{Total-Text}} & \multirow{2}{*}{\textbf{FPS}} & \multirow{2}{*}{\textbf{VRAM}}\\
    & P & R & F & & \\
    \midrule[1.1pt]
    DPText-DETR & - & - & 86.79 & 13.6 & 17409MB\\
    \hline
    Box2Poly(ours) & 89.58 & 83.83 & 86.60 & \textbf{19.6} & \textbf{8072MB}\\
    \bottomrule[1.5pt]
    \end{tabular}
    }
    \caption{Qualitative results on TotalText without pretrainig. VRAM occupation is measured with 2 images per GPU.}
    \label{tab:no_pretrain_results}
\end{table}

\subsubsection{Polygon IoU Loss}
Since its inception, the Intersection over Union (IoU) loss \cite{yu2016unitbox} has emerged as an indispensable technique in the realm of bounding box object detection. Therefore, we are also inclined to devise a similar loss to aid the task of polygon prediction. Recognizing that formulating the intersection between two non-convex polygons in a differentiable manner is a challenging endeavor, we address this issue by decomposing the polygon into a set of quadrilaterals. The IoU loss is then calculated at this level. Notably, this polygon IoU loss does not yield any improvement, and even leads to a performance drop (as evidenced by Exp. 5 in Tab.~\ref{tab:ablations}). We assume the reason behind this is, unlike bounding box, IoU loss is redundant when vertex-level \(L1\) loss is introduced, and this estimated IoU cannot reflect the overlapping quality between non-convex polygons well.
\section{Conclusion}
In this paper, we present a novel approach for text detection by performing iterative polygon regression. Our method modifies the box iterative regression pipeline to make it suit the polygon regression task. It begins with a learnable bounding box and incorporates a transformative process, allowing for seamless transition from box to polygon representation. Then, the polygon coordinates are iteratively refined using polygon RoI(Region of Interest) features. Through this process, our network demonstrates remarkable memory efficiency and stability in detecting text of varying and intricate shapes. Despite a slight reduction in performance compared to the current state-of-the-art (SOTA), our approach still maintains commendable results.  It's worth noting that our detector may face challenges in scenarios where small texts dominate the scene, difficult to guarantee high recall. To address this, the incorporation of a reliable proposal generator could be beneficial. Additionally, any resulting increase in memory occupation due to this augmentation can be probably counterbalanced by reducing the number of box and polygon regression layers.

\bibliography{references.bib}

\begin{thebibliography}{44}
\providecommand{\natexlab}[1]{#1}

\bibitem[{Baek et~al.(2019)Baek, Lee, Han, Yun, and Lee}]{baek2019character}
Baek, Y.; Lee, B.; Han, D.; Yun, S.; and Lee, H. 2019.
\newblock Character region awareness for text detection.
\newblock In \emph{Proceedings of the IEEE/CVF conference on computer vision
  and pattern recognition}, 9365--9374.

\bibitem[{Bi and Hu(2021)}]{bi2021disentangled}
Bi, Y.; and Hu, Z. 2021.
\newblock Disentangled contour learning for quadrilateral text detection.
\newblock In \emph{Proceedings of the IEEE/CVF Winter Conference on
  Applications of Computer Vision}, 909--918.

\bibitem[{Cai and Vasconcelos(2018)}]{cai2018cascade}
Cai, Z.; and Vasconcelos, N. 2018.
\newblock Cascade r-cnn: Delving into high quality object detection.
\newblock In \emph{Proceedings of the IEEE conference on computer vision and
  pattern recognition}, 6154--6162.

\bibitem[{Carion et~al.(2020)Carion, Massa, Synnaeve, Usunier, Kirillov, and
  Zagoruyko}]{carion2020end}
Carion, N.; Massa, F.; Synnaeve, G.; Usunier, N.; Kirillov, A.; and Zagoruyko,
  S. 2020.
\newblock End-to-end object detection with transformers.
\newblock In \emph{European conference on computer vision}, 213--229. Springer.

\bibitem[{Ch'ng and Chan(2017)}]{ch2017total}
Ch'ng, C.~K.; and Chan, C.~S. 2017.
\newblock Total-text: A comprehensive dataset for scene text detection and
  recognition.
\newblock In \emph{2017 14th IAPR international conference on document analysis
  and recognition (ICDAR)}, volume~1, 935--942. IEEE.

\bibitem[{Dai et~al.(2021{\natexlab{a}})Dai, Zhang, Zhang, and
  Cao}]{dai2021progressive}
Dai, P.; Zhang, S.; Zhang, H.; and Cao, X. 2021{\natexlab{a}}.
\newblock Progressive contour regression for arbitrary-shape scene text
  detection.
\newblock In \emph{Proceedings of the IEEE/CVF conference on computer vision
  and pattern recognition}, 7393--7402.

\bibitem[{Dai et~al.(2021{\natexlab{b}})Dai, Chen, Yang, Zhang, Yuan, and
  Zhang}]{dai2021dynamic}
Dai, X.; Chen, Y.; Yang, J.; Zhang, P.; Yuan, L.; and Zhang, L.
  2021{\natexlab{b}}.
\newblock Dynamic detr: End-to-end object detection with dynamic attention.
\newblock In \emph{Proceedings of the IEEE/CVF International Conference on
  Computer Vision}, 2988--2997.

\bibitem[{Du et~al.(2022)Du, Ye, Zhang, Liu, and Tao}]{du2022i3cl}
Du, B.; Ye, J.; Zhang, J.; Liu, J.; and Tao, D. 2022.
\newblock I3cl: Intra-and inter-instance collaborative learning for
  arbitrary-shaped scene text detection.
\newblock \emph{International Journal of Computer Vision}, 130(8): 1961--1977.

\bibitem[{He et~al.(2017)He, Gkioxari, Doll{\'a}r, and Girshick}]{he2017mask}
He, K.; Gkioxari, G.; Doll{\'a}r, P.; and Girshick, R. 2017.
\newblock Mask r-cnn.
\newblock In \emph{Proceedings of the IEEE international conference on computer
  vision}, 2961--2969.

\bibitem[{He et~al.(2016)He, Zhang, Ren, and Sun}]{he2016deep}
He, K.; Zhang, X.; Ren, S.; and Sun, J. 2016.
\newblock Deep residual learning for image recognition.
\newblock In \emph{Proceedings of the IEEE conference on computer vision and
  pattern recognition}, 770--778.

\bibitem[{He et~al.(2021)He, Liao, Yang, Zhong, Tang, Cheng, Yao, Wang, and
  Bai}]{he2021most}
He, M.; Liao, M.; Yang, Z.; Zhong, H.; Tang, J.; Cheng, W.; Yao, C.; Wang, Y.;
  and Bai, X. 2021.
\newblock MOST: A multi-oriented scene text detector with localization
  refinement.
\newblock In \emph{Proceedings of the IEEE/CVF Conference on Computer Vision
  and Pattern Recognition}, 8813--8822.

\bibitem[{Hong et~al.(2019)Hong, Petillot, Lane, Miao, and
  Wang}]{hong2019textplace}
Hong, Z.; Petillot, Y.; Lane, D.; Miao, Y.; and Wang, S. 2019.
\newblock TextPlace: Visual place recognition and topological localization
  through reading scene texts.
\newblock In \emph{Proceedings of the IEEE/CVF International Conference on
  Computer Vision}, 2861--2870.

\bibitem[{Huang et~al.(2022)Huang, Liu, Peng, Liu, Lin, Zhu, Yuan, Ding, and
  Jin}]{huang2022swintextspotter}
Huang, M.; Liu, Y.; Peng, Z.; Liu, C.; Lin, D.; Zhu, S.; Yuan, N.; Ding, K.;
  and Jin, L. 2022.
\newblock Swintextspotter: Scene text spotting via better synergy between text
  detection and text recognition.
\newblock In \emph{proceedings of the IEEE/CVF conference on computer vision
  and pattern recognition}, 4593--4603.

\bibitem[{Li et~al.(2020)Li, Zou, Sartori, Pei, and Yu}]{li2020textslam}
Li, B.; Zou, D.; Sartori, D.; Pei, L.; and Yu, W. 2020.
\newblock Textslam: Visual slam with planar text features.
\newblock In \emph{2020 IEEE International Conference on Robotics and
  Automation (ICRA)}, 2102--2108. IEEE.

\bibitem[{Liao et~al.(2022)Liao, Chen, Wang, Cheng, Zhang, Liu, and
  Huang}]{liao2022maptr}
Liao, B.; Chen, S.; Wang, X.; Cheng, T.; Zhang, Q.; Liu, W.; and Huang, C.
  2022.
\newblock MapTR: Structured Modeling and Learning for Online Vectorized HD Map
  Construction.
\newblock \emph{arXiv preprint arXiv:2208.14437}.

\bibitem[{Liao et~al.(2020)Liao, Wan, Yao, Chen, and Bai}]{liao2020real}
Liao, M.; Wan, Z.; Yao, C.; Chen, K.; and Bai, X. 2020.
\newblock Real-time scene text detection with differentiable binarization.
\newblock In \emph{Proceedings of the AAAI conference on artificial
  intelligence}, volume~34, 11474--11481.

\bibitem[{Lin et~al.(2017)Lin, Goyal, Girshick, He, and
  Doll{\'a}r}]{lin2017focal}
Lin, T.-Y.; Goyal, P.; Girshick, R.; He, K.; and Doll{\'a}r, P. 2017.
\newblock Focal loss for dense object detection.
\newblock In \emph{Proceedings of the IEEE international conference on computer
  vision}, 2980--2988.

\bibitem[{Liu et~al.(2020)Liu, Chen, Shen, He, Jin, and Wang}]{liu2020abcnet}
Liu, Y.; Chen, H.; Shen, C.; He, T.; Jin, L.; and Wang, L. 2020.
\newblock Abcnet: Real-time scene text spotting with adaptive bezier-curve
  network.
\newblock In \emph{proceedings of the IEEE/CVF conference on computer vision
  and pattern recognition}, 9809--9818.

\bibitem[{Liu and Jin(2017)}]{liu2017deep}
Liu, Y.; and Jin, L. 2017.
\newblock Deep matching prior network: Toward tighter multi-oriented text
  detection.
\newblock In \emph{Proceedings of the IEEE conference on computer vision and
  pattern recognition}, 1962--1969.

\bibitem[{Liu et~al.(2021)Liu, Shen, Jin, He, Chen, Liu, and
  Chen}]{liu2021abcnet}
Liu, Y.; Shen, C.; Jin, L.; He, T.; Chen, P.; Liu, C.; and Chen, H. 2021.
\newblock Abcnet v2: Adaptive bezier-curve network for real-time end-to-end
  text spotting.
\newblock \emph{IEEE Transactions on Pattern Analysis and Machine
  Intelligence}, 44(11): 8048--8064.

\bibitem[{Long et~al.(2018)Long, Ruan, Zhang, He, Wu, and
  Yao}]{long2018textsnake}
Long, S.; Ruan, J.; Zhang, W.; He, X.; Wu, W.; and Yao, C. 2018.
\newblock Textsnake: A flexible representation for detecting text of arbitrary
  shapes.
\newblock In \emph{Proceedings of the European conference on computer vision
  (ECCV)}, 20--36.

\bibitem[{Lorentz(2012)}]{lorentz2012bernstein}
Lorentz, G.~G. 2012.
\newblock \emph{Bernstein polynomials}.
\newblock American Mathematical Soc.

\bibitem[{Lyu et~al.(2018)Lyu, Liao, Yao, Wu, and Bai}]{lyu2018mask}
Lyu, P.; Liao, M.; Yao, C.; Wu, W.; and Bai, X. 2018.
\newblock Mask textspotter: An end-to-end trainable neural network for spotting
  text with arbitrary shapes.
\newblock In \emph{Proceedings of the European conference on computer vision
  (ECCV)}, 67--83.

\bibitem[{Ma et~al.(2018)Ma, Shao, Ye, Wang, Wang, Zheng, and
  Xue}]{ma2018arbitrary}
Ma, J.; Shao, W.; Ye, H.; Wang, L.; Wang, H.; Zheng, Y.; and Xue, X. 2018.
\newblock Arbitrary-oriented scene text detection via rotation proposals.
\newblock \emph{IEEE transactions on multimedia}, 20(11): 3111--3122.

\bibitem[{Nayef et~al.(2019)Nayef, Patel, Busta, Chowdhury, Karatzas, Khlif,
  Matas, Pal, Burie, Liu et~al.}]{nayef2019icdar2019}
Nayef, N.; Patel, Y.; Busta, M.; Chowdhury, P.~N.; Karatzas, D.; Khlif, W.;
  Matas, J.; Pal, U.; Burie, J.-C.; Liu, C.-l.; et~al. 2019.
\newblock ICDAR2019 robust reading challenge on multi-lingual scene text
  detection and recognition—RRC-MLT-2019.
\newblock In \emph{2019 International conference on document analysis and
  recognition (ICDAR)}, 1582--1587. IEEE.

\bibitem[{Ren et~al.(2015)Ren, He, Girshick, and Sun}]{ren2015faster}
Ren, S.; He, K.; Girshick, R.; and Sun, J. 2015.
\newblock Faster r-cnn: Towards real-time object detection with region proposal
  networks.
\newblock \emph{Advances in neural information processing systems}, 28.

\bibitem[{Rong et~al.(2016)Rong, Li, Munoz, Xiao, Arditi, and
  Tian}]{rong2016guided}
Rong, X.; Li, B.; Munoz, J.~P.; Xiao, J.; Arditi, A.; and Tian, Y. 2016.
\newblock Guided text spotting for assistive blind navigation in unfamiliar
  indoor environments.
\newblock In \emph{Advances in Visual Computing: 12th International Symposium,
  ISVC 2016, Las Vegas, NV, USA, December 12-14, 2016, Proceedings, Part II
  12}, 11--22. Springer.

\bibitem[{Sun and Liu(2022)}]{sun2022center}
Sun, L.; and Liu, K. 2022.
\newblock Center TextSpotter: A novel text spotter for autonomous unmanned
  vehicles.
\newblock \emph{IEEE Transactions on Intelligent Transportation Systems}.

\bibitem[{Sun et~al.(2021)Sun, Zhang, Jiang, Kong, Xu, Zhan, Tomizuka, Li,
  Yuan, Wang et~al.}]{sun2021sparse}
Sun, P.; Zhang, R.; Jiang, Y.; Kong, T.; Xu, C.; Zhan, W.; Tomizuka, M.; Li,
  L.; Yuan, Z.; Wang, C.; et~al. 2021.
\newblock Sparse r-cnn: End-to-end object detection with learnable proposals.
\newblock In \emph{Proceedings of the IEEE/CVF conference on computer vision
  and pattern recognition}, 14454--14463.

\bibitem[{Wang et~al.(2020)Wang, Lu, Zhang, Yang, Bai, Xu, He, Wang, and
  Liu}]{wang2020all}
Wang, H.; Lu, P.; Zhang, H.; Yang, M.; Bai, X.; Xu, Y.; He, M.; Wang, Y.; and
  Liu, W. 2020.
\newblock All you need is boundary: Toward arbitrary-shaped text spotting.
\newblock In \emph{Proceedings of the AAAI conference on artificial
  intelligence}, volume~34, 12160--12167.

\bibitem[{Wang et~al.(2019)Wang, Xie, Song, Zang, Wang, Lu, Yu, and
  Shen}]{wang2019efficient}
Wang, W.; Xie, E.; Song, X.; Zang, Y.; Wang, W.; Lu, T.; Yu, G.; and Shen, C.
  2019.
\newblock Efficient and accurate arbitrary-shaped text detection with pixel
  aggregation network.
\newblock In \emph{Proceedings of the IEEE/CVF international conference on
  computer vision}, 8440--8449.

\bibitem[{Xu et~al.(2019)Xu, Wang, Zhou, Wang, Yang, and Bai}]{xu2019textfield}
Xu, Y.; Wang, Y.; Zhou, W.; Wang, Y.; Yang, Z.; and Bai, X. 2019.
\newblock Textfield: Learning a deep direction field for irregular scene text
  detection.
\newblock \emph{IEEE Transactions on Image Processing}, 28(11): 5566--5579.

\bibitem[{Yang et~al.(2021{\natexlab{a}})Yang, Yan, Ming, Wang, Zhang, and
  Tian}]{yang2021rethinking}
Yang, X.; Yan, J.; Ming, Q.; Wang, W.; Zhang, X.; and Tian, Q.
  2021{\natexlab{a}}.
\newblock Rethinking rotated object detection with gaussian wasserstein
  distance loss.
\newblock In \emph{International conference on machine learning}, 11830--11841.
  PMLR.

\bibitem[{Yang et~al.(2021{\natexlab{b}})Yang, Yang, Yang, Ming, Wang, Tian,
  and Yan}]{yang2021learning}
Yang, X.; Yang, X.; Yang, J.; Ming, Q.; Wang, W.; Tian, Q.; and Yan, J.
  2021{\natexlab{b}}.
\newblock Learning high-precision bounding box for rotated object detection via
  kullback-leibler divergence.
\newblock \emph{Advances in Neural Information Processing Systems}, 34:
  18381--18394.

\bibitem[{Yang et~al.(2022)Yang, Zhou, Zhang, Yang, Wang, Yan, Zhang, and
  Tian}]{yang2022kfiou}
Yang, X.; Zhou, Y.; Zhang, G.; Yang, J.; Wang, W.; Yan, J.; Zhang, X.; and
  Tian, Q. 2022.
\newblock The KFIoU loss for rotated object detection.
\newblock \emph{arXiv preprint arXiv:2201.12558}.

\bibitem[{Ye et~al.(2023{\natexlab{a}})Ye, Zhang, Zhao, Liu, Du, and
  Tao}]{ye2023dptext}
Ye, M.; Zhang, J.; Zhao, S.; Liu, J.; Du, B.; and Tao, D. 2023{\natexlab{a}}.
\newblock Dptext-detr: Towards better scene text detection with dynamic points
  in transformer.
\newblock In \emph{Proceedings of the AAAI Conference on Artificial
  Intelligence}, volume~37, 3241--3249.

\bibitem[{Ye et~al.(2023{\natexlab{b}})Ye, Zhang, Zhao, Liu, Liu, Du, and
  Tao}]{ye2023deepsolo}
Ye, M.; Zhang, J.; Zhao, S.; Liu, J.; Liu, T.; Du, B.; and Tao, D.
  2023{\natexlab{b}}.
\newblock Deepsolo: Let transformer decoder with explicit points solo for text
  spotting.
\newblock In \emph{Proceedings of the IEEE/CVF Conference on Computer Vision
  and Pattern Recognition}, 19348--19357.

\bibitem[{Yu et~al.(2016)Yu, Jiang, Wang, Cao, and Huang}]{yu2016unitbox}
Yu, J.; Jiang, Y.; Wang, Z.; Cao, Z.; and Huang, T. 2016.
\newblock Unitbox: An advanced object detection network.
\newblock In \emph{Proceedings of the 24th ACM international conference on
  Multimedia}, 516--520.

\bibitem[{Zhang et~al.(2019)Zhang, Liang, Huang, En, Han, Ding, and
  Ding}]{zhang2019look}
Zhang, C.; Liang, B.; Huang, Z.; En, M.; Han, J.; Ding, E.; and Ding, X. 2019.
\newblock Look more than once: An accurate detector for text of arbitrary
  shapes.
\newblock In \emph{Proceedings of the IEEE/CVF conference on computer vision
  and pattern recognition}, 10552--10561.

\bibitem[{Zhang et~al.(2023)Zhang, Yang, Zhu, and Yin}]{zhang2023arbitrary}
Zhang, S.-X.; Yang, C.; Zhu, X.; and Yin, X.-C. 2023.
\newblock Arbitrary shape text detection via boundary transformer.
\newblock \emph{IEEE Transactions on Multimedia}.

\bibitem[{Zhang et~al.(2022)Zhang, Su, Tripathi, and Tu}]{zhang2022text}
Zhang, X.; Su, Y.; Tripathi, S.; and Tu, Z. 2022.
\newblock Text spotting transformers.
\newblock In \emph{Proceedings of the IEEE/CVF Conference on Computer Vision
  and Pattern Recognition}, 9519--9528.

\bibitem[{Zhang et~al.(2016)Zhang, Zhang, Shen, Yao, Liu, and
  Bai}]{zhang2016multi}
Zhang, Z.; Zhang, C.; Shen, W.; Yao, C.; Liu, W.; and Bai, X. 2016.
\newblock Multi-oriented text detection with fully convolutional networks.
\newblock In \emph{Proceedings of the IEEE conference on computer vision and
  pattern recognition}, 4159--4167.

\bibitem[{Zhu et~al.(2020)Zhu, Su, Lu, Li, Wang, and Dai}]{zhu2020deformable}
Zhu, X.; Su, W.; Lu, L.; Li, B.; Wang, X.; and Dai, J. 2020.
\newblock Deformable detr: Deformable transformers for end-to-end object
  detection.
\newblock \emph{arXiv preprint arXiv:2010.04159}.

\bibitem[{Zhu et~al.(2017)Zhu, Liao, Yang, and Liu}]{zhu2017cascaded}
Zhu, Y.; Liao, M.; Yang, M.; and Liu, W. 2017.
\newblock Cascaded segmentation-detection networks for text-based traffic sign
  detection.
\newblock \emph{IEEE transactions on intelligent transportation systems},
  19(1): 209--219.

\end{thebibliography}

\newpage

\section{Appendix}
\subsection{A. Different Implementations of PolyAlign}

\begin{algorithm}
\caption{The Algorithm of RoIAlign}\label{alg:roi_align}
\begin{algorithmic}
\REQUIRE 
\STATE Coordinates of IoU region;
\STATE $f$: feature map generated from input image;
\STATE $sampling$: number of sampling points;
\STATE $out_{w}$: the output width of RoIAlign;
\STATE $out_{h}$: the output height of RoIAlign;
\ENSURE
\STATE The extracted RoI feature: \\$result = \left[result_{i,j}\right]_{i=0,j=0}^{out_{w}-1,out_{h}-1}$
\FOR{$i=0;i\leq out_{w}-1;i{+}{+}$}
    \FOR{$j=0;j\leq out_{h}-1;j{+}{+}$}
    \STATE $result_{i,j} \gets 0$
    \FOR{$ii=0;ii\leq sampling-1;ii{+}{+}$}
    \FOR{$jj=0;jj\leq sampling-1;jj{+}{+}$}
    \STATE $r \gets bilinear\textunderscore interpolate\left(f, x_{ii}, y_{jj}\right)$
    \STATE $\backslash \backslash \left(x_{ii}, y_{jj}\right)$ represents the coordinates of the corresponding sampling point.
    \STATE $result_{i,j} \gets result_{i,j} + r$
    \ENDFOR
    \ENDFOR
    \STATE $result_{i,j} \gets result_{i,j}/sampling^2 $
    \ENDFOR
\ENDFOR
\end{algorithmic}
\end{algorithm}

\begin{figure}[h]
	\centering
	\includegraphics[width=0.35\textwidth]{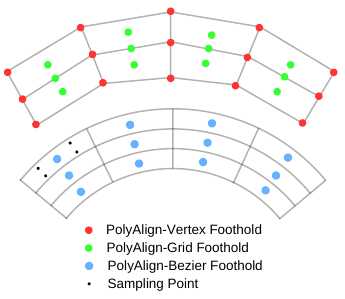}
	\caption{The illustration of feature extracting locations of PolyAlign-Vertex \& PolyAlign-Grid (top) and PolyAlign-Bezier(bottom)}
	\label{fig:polyalign_new}
\end{figure}

The pseudocode for RoIAlign is presented in Algorithm~\ref{alg:roi_align}. The implementation of PolyAlign follows a same mechanism, with the primary distinction lying in the locations where bilinear interpolation is computed.

\begin{figure}[h]
	\centering
	\includegraphics[width=0.4\textwidth]{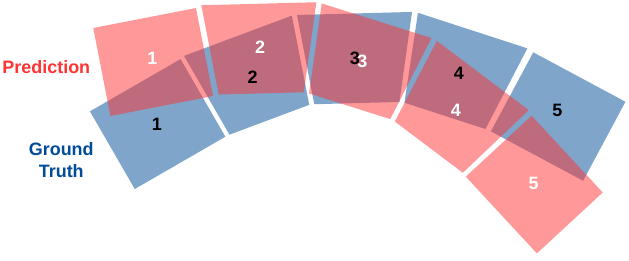}
	\caption{The illustration of calculating poly IoU loss between polygons. prediction and ground truth polygons are decomposed to quadrilaterals.}
	\label{fig:poly_iou}
\end{figure}
\subsubsection{PolyAlign-Vertex \& PolyAlign-Grid}
As depicted in Figure~\ref{fig:polyalign_new} (top), PolyAlign-Vertex operates directly on the vertices of both polygons and centerlines. In contrast, PolyAlign-Grid performs bilinear interpolation at the center of each grid within the polygon RoI. Both implementations utilize a sampling point count of 1.
\begin{figure}[h]
	\centering
	\includegraphics[width=0.48\textwidth]{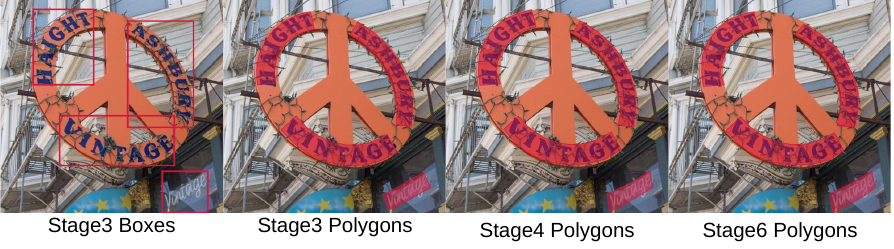}
	\caption{Visualization of predicted boxes/polygons of several stages in iterative regression. "Stage3 Polygons" means the polygons are generated by box-to-polygon transformation.}
	\label{fig:regression_process}
\end{figure}

\begin{figure*}[!ht]
	\centering
	\includegraphics[width=0.93\textwidth]{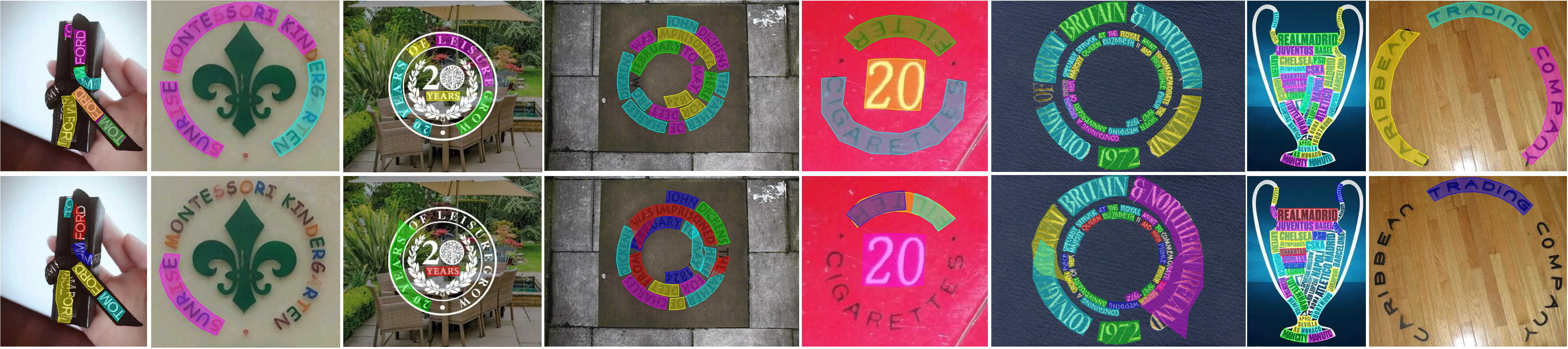}
	\caption{Visual comparison of Box2Poly(top row) and DPText-DETR(bottom row). Please zoom in to as needed.}
	\label{fig:visual_compare}
\end{figure*}
\subsubsection{PolyAlign-Bezier}
The PolyAlign-Bezier is implemented based on BezierAlign \cite{liu2020abcnet}, but several modifications are made: Firstly, a bug related to the computation of feature extraction footholds (blue points in Fig.~\ref{fig:polyalign_new}(bottom)) in BezierAlign has been rectified. The bilinear interpolation is then conducted around uniformly distributed grid center. Second, the sampling points around each grid center are arranged following the text orientation instead of horizontally. A count of 2 sampling points is employed for each RoI grid.
\subsection{B. Polygon IoU Loss}
As illustrated in Figure~\ref{fig:poly_iou}, both the predicted and ground truth polygons are disassembled into a series of quadrilaterals by slicing them through corresponding top and bottom vertices. Subsequently, the Intersection over Union (IoU) is computed by assessing the overlap between the paired quadrilaterals, each identified by a matching id number.

Our implementation of quadrilateral IoU is built upon the work available at \url{https://github.com/lilanxiao/Rotated_IoU}, albeit with several adaptations. The area of each quadrilateral is determined utilizing the shoelace formula. Moreover, the methodology to calculate the intersection between two rotated bounding boxes has been revised to cater specifically to quadrilaterals, accommodating their unique characteristics.

\subsection{C. More Visual Results}
\begin{figure}[h]
	\centering
	\includegraphics[width=0.48\textwidth]{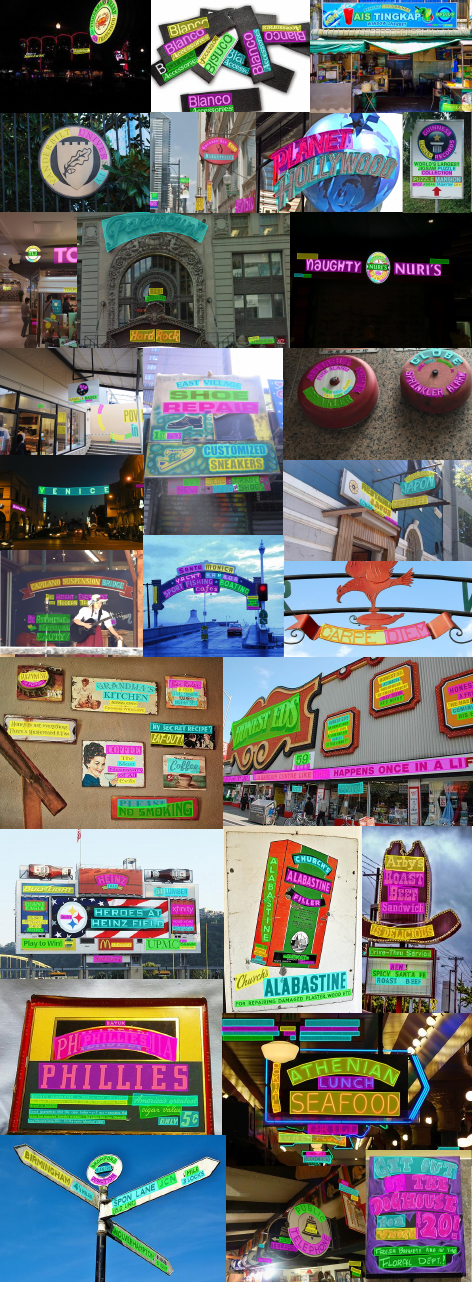}
	\caption{Visual results of corner cases. Please zoom in as required}
	\label{fig:visual_result}
\end{figure}

In comparison with other polygon-based methods, the main differences occur on non-horizontal and curved text. As shown in Fig.\ref{fig:visual_compare}, DPText-DETR~\cite{ye2023dptext} struggles to precisely delineate such text, whereas Box2Poly retrieves tighter, well-aligned boundaries. We also visualize the results of our method for several corner cases from the test sets of TotalText and CTW1500 in Fig.\ref{fig:visual_result}.

We note that the techniques included in the paper are not restricted to Sparse RCNN, but can be combined with any bounding box detector. They are potentially useful building blocks for effective, lightweight text detection, beyond our specific implementation. Figure \ref{fig:regression_process} illustrates the gradual enhancement of the text detection result: given bounding box priors from previous stage, the box-to-polygon transformation already finds fairly tight outlines for each text instance, still extra regression steps (stages 4 and 6) visibly improve the fit. If the reviewer finds our explanations convincing we kindly ask them to reconsider their rating.

\subsection{D. Efficiency of Our Method}
The faster inference speed is due mainly to the simpler feature embedding compared to DPText-DETR (1 vs.\ 16 embeddings/instances). The Box2Poly configuration used to obtain the final results in the paper has $\approx$170 GFLOPs, similar to Deformable DETR (173 GFLOPs). Since DPText-DETR uses its own implementation of Deformable Attention we would need more time to work out the exact FLOP count, but it must definitely be higher than Deformable DETR because of the point-level feature embedding and correspondingly more cross-attentions and self-attentions.
We should say that our current implementation of PolyAlign is not optimised for performance yet, we simply build it on Pytorch's built-in function \emph{torch.nn.functional.grid\_sample}. We believe that a CUDA implementation could make Box2Poly as efficient as Sparse RCNN($\sim$130 GFLOPs) as one could largely follow the design of RoIAlign in terms of memory access and interpolation.

\end{document}